\documentclass[12pt, oneside]{article}
\usepackage{graphicx,color}
\usepackage{mdframed}
\usepackage[utf8]{inputenc}

\usepackage{amsmath}
\usepackage{amsfonts}
\usepackage{amssymb}
\usepackage{amsthm}

\usepackage{newtxtext}
\usepackage{newtxmath}

\usepackage{tabularx}
\usepackage{booktabs}

\usepackage[authoryear,sort]{natbib}
\usepackage{amssymb}
\usepackage{amsthm}
\usepackage{float}
\usepackage{enumitem}
\usepackage[svgnames,table]{xcolor}
\usepackage{hyperref}
\hypersetup{
  colorlinks=true,
  allcolors=blue
}

\usepackage{authblk}
\usepackage{comment}
\usepackage{booktabs}
\usepackage{multirow}
\usepackage{xurl}
\usepackage[most]{tcolorbox}
\usepackage{fvextra}

\definecolor{hcolor}{HTML}{7FBAB0}

\newcommand{\keywords}[1]{\textbf{Keywords:} #1}

\newtcolorbox{promptbox}{
  enhanced,
  colback=gray!3,
  colframe=gray!60,
  boxrule=0.5pt,
  arc=2pt,
  left=6pt,right=6pt,top=6pt,bottom=6pt,
  breakable,
  title=System Prompt,
  fonttitle=\small\bfseries
}

\DefineVerbatimEnvironment{promptverbatim}{Verbatim}{
  fontsize=\small,
  breaklines=true,
  breakanywhere=true,
  breaksymbolleft={},
  commandchars=\\\{\}
}

\newcounter{prompt}

\begin{document}

\title{Can You Trust What You See? Human and AI Detection of Synthetic Legal Evidence}

\author[1]{Jinzhe Tan\thanks{Corresponding author. Email: jinzhe.tan@umontreal.ca}}
\author[1,2]{Ali Ekber Cinar}
\author[1]{Karim Benyekhlef}

\affil[1]{Cyberjustice Laboratory, Faculty of Law, Universit\'{e} de Montr\'{e}al}
\affil[2]{Faculty of Law, McGill University}

\date{}

\maketitle
\vspace{-7mm}

\begin{abstract}
\noindent 
Visual evidence has long been treated as a reliable form of legal proof, but advances in artificial intelligence (AI) are undermining that assumption. This article asks how well humans and frontier multimodal large language models (MLLMs) can distinguish authentic evidentiary photographs from AI-generated counterparts in the object-centric scenarios typical of civil disputes. We built \textit{Synthetic Legal Evidence Detection} (SLED-1400), a dataset of 200 authentic evidence images paired with 1,200 synthetic counterparts produced by six contemporary text-to-image generators across ten evidence categories. The same stimuli and response format were used in a controlled web experiment with 136 lay participants and in a standardized evaluation of four MLLMs (GPT-5.1, Gemini-3-Pro, Gemini-3-Flash, Qwen3-VL-235B). Human accuracy was 64.8\,\% overall, and 48.5\,\% and 51.0\,\% on the two strongest generators (Gemini-3-Pro-Image and Flux-2-Max), indistinguishable from chance. MLLMs never misclassified an authentic image (100\,\% specificity), but missed most synthetic outputs from the harder generators, with average MLLM detection at 5.9\,\% on Gemini-3-Pro-Image outputs. Human and MLLM errors were largely uncorrelated, while the four MLLMs were strongly correlated with each other. Neither group is a reliable standalone authenticator. We argue that visual evidence in legal proceedings should be treated as inherently contestable, and that a workable procedural response must combine trained human review, MLLM screening, and provenance infrastructure such as C2PA Content Credentials.

\vspace{6mm}
\noindent\keywords{AI-generated images; visual evidence; synthetic media detection; legal AI; multimodal large language models; evidence authentication}
\end{abstract}

\section{Introduction}

Digital images have become a routine form of ``everyday evidence'' in contemporary adjudication and everyday dispute \citep{golan2008visual}. Litigants, consumers, merchants, insurers, and platform moderators regularly rely on photos of damaged packages, defective goods, property loss, or incident scenes to support factual claims. These decisions are often made under time pressure and asymmetric information, as decision-makers rarely have access to the physical object and must infer liability from visual evidence and short narratives.

Recent advances in generative artificial intelligence (AI) are challenging this long-standing assumption of \textit{prima facie} trust in visual evidence. Early ``deepfake'' techniques were primarily associated with face swapping \citep{gerstner_faceoff_2020,waseem_deepfake_2023} or audio impersonation \citep{govindaraju_manipulating_2023, yan_deep_2023}. Consequently, they attracted attention for identity-related harms \citep{govindaraju_manipulating_2023, yan_deep_2023} and large scale misinformation \cite{chesney2019deep}. By contrast, contemporary MLLMs enable semantic-level forgery. They can fabricate visual evidence that never existed or manipulate authentic images to alter their evidentiary meaning. As these models improve their rendering of lighting, material properties, and texture continuity, the synthetic images can remain visually coherent with the surrounding scene while directly targeting the fact in dispute.

Real-world disputes have already begun to confront the consequences of synthetic evidence, especially in ``high-volume, low-value'' settings \citep{benyekhlef2014low}. In its reporting on refund-only scams, \cite{Li2025AIRefundFraud} describes cases in which customers allegedly submitted AI-altered images of rotten or damaged goods to obtain refunds, while platforms and service centers indicated that they could not determine whether the images had been AI-generated. Concerns are not limited to everyday disputes. In one of the earliest documented judicial instances, a California court rejected and sanctioned the submission of AI-generated video evidence that purported to depict a real witness \citep{perlo_ai-generated_2025}. These cases illustrate that the evidentiary reliability of user-submitted visual content can no longer be taken for granted.

The implications extend beyond isolated incidents. In civil litigation, fabricated images of product defects or service failures could support illegitimate claims for compensation. In criminal proceedings, synthetic images could be used to construct false alibis or to incriminate innocent parties. The challenge is compounded by the rapid pace of technological advancement. Each new generation of MLLMs produces outputs that are more realistic and harder to detect than the last \citep{karras2019style,ramesh2022hierarchical,rombach2022high}. As a result, visual and textual evidence that has long been treated as relatively trustworthy now raises acute challenges for evidentiary standards, due process, and institutional safeguards in both formal courts and high-volume dispute resolution systems.

Despite growing concern, significant gaps remain in our empirical understanding of how well humans and technical tools can detect AI-generated visual evidence in legal contexts. Most prior work focuses on identity-centric manipulations (e.g., faces, voices) or broad misinformation settings \citep{dolhansky2020deepfake,rossler2019faceforensics++,livernoche2025openfake}. Legal and everyday disputes, however, are typically object-centric. The relevant factual question often turns on small, localized features such as cracks, stains or leaks. Moreover, as generative models continue to evolve, previously reliable cues for detecting synthetic imagery (e.g., distorted text, or visible watermarks) are becoming increasingly unstable, suggesting that both human and automated detection systems face a moving target.

This paper addresses these gaps by studying how well humans and state-of-the-art MLLMs can distinguish authentic visual evidence from AI-generated counterparts in legal settings. We propose three research questions (RQ): 

\textbf{RQ1.} How accurately can humans identify AI-synthetic legal evidence images?

\textbf{RQ2.} How do \emph{MLLMs} perform compared to humans in identifying AI-synthetic evidence images?

\textbf{RQ3.} How does detectability vary across image generation models and evidence categories?

The main contributions of this paper are as follows: \textbf{(1)} We construct a paired dataset of authentic and AI-synthetic images grounded in realistic legal and online dispute resolution (ODR) scenarios. \textbf{(2)} We conduct a controlled human study with confidence and reasoning annotations and benchmark performance against multiple state-of-the-art MLLMs under a unified evaluation setting. \textbf{(3)} We identify substantial variation in detectability across generation models and document a pronounced conservative bias in current MLLMs, with direct implications for evidence authentication and AI-assisted verification.

The remainder of this article is structured as follows. Section~\ref{sec:related-work} reviews related work on AI in legal contexts, synthetic media perception, and detection methods. Section~\ref{sec:method} presents the dataset construction and experimental design. Section~\ref{sec:results} reports results from human and MLLM evaluations. Section~\ref{sec:discussion} discusses implications for legal evidence and verification systems. Section~\ref{sec:limitations} discusses the limitations of our study. Finally, Section~\ref{sec:conclusion} concludes with broader reflections and future directions.

\section{Related Work}
\label{sec:related-work}

\subsection{AI in Legal Contexts}

The use of AI in legal settings has grown significantly over time. Early research focused on relatively narrow tasks, such as improving legal information retrieval, extracting relevant information from statutes and regulations, and identifying argumentative elements in judicial decisions \citep{ashley_artificial_2017,bucher_navigating_2025}. More recent developments, however, have expanded both the range and complexity of AI applications across legal domains.

In particular, generative AI has begun to reshape aspects of legal practice. It is increasingly used to support access to justice \citep{westermann2023bridging}, dispute resolution \citep{tan2024robots}, and decision support. Most current applications focus on text-heavy tasks, including legal research, document drafting, summarization, and classification. These uses can improve efficiency and, in some cases, expand access to legal help \citep{tan2023chatgpt}. 

Beyond practice-oriented tasks, generative AI is also being explored as a tool for legal education \citep{cinar_language_2024} and for supporting regulatory compliance \citep{turksen_legal_2024}. 
At the same time, advances in multimodal AI systems, which can process not only text but also images and video, are extending these capabilities further \citep{qiang2025ver}. Emerging work explores the use of such models for document review \citep{westermann2024analyzing}, understanding legal websites \citep{tan2025legalwebagent}, and, more directly, for analyzing evidentiary materials or legally relevant visual records \citep{kim2025digital,hoeben2025can}. 

\subsection{Human Perception of Synthetic Media}

While the rapidly advancing technologies that enable the creation of synthetic media introduce a range of substantial benefits, they also bring major risks related to human perception, including deception and the erosion of trust in audiovisual information \citep{barraclough_perception_2019}. These challenges are particularly salient in the context of shifting perceptions of reality. Synthetic media increasingly renders reality malleable, personalized, and susceptible to manipulation. As a result, some scholars have argued that reality itself may need to be reconceptualized in terms of its affective impact on individuals, given that synthetic experiences can meaningfully influence beliefs, decisions, and behavior despite lacking a physical basis \citep{kalpokas_problematising_2020}.

The consequences of these changing perceptions are especially pronounced in judicial contexts. The literature consistently emphasizes that the proliferation of inauthentic visual media undermines epistemic trust in images and audiovisual materials within legal settings \citep{murray_visual_2025}. This erosion of trust poses a significant challenge for the judicial system, as the introduction of deepfakes into legal proceedings not only destabilizes confidence in audiovisual evidence but also places strain on established standards and practices of authentication \citep{mullen_new_2022}.

\subsection{Human Detection of AI-Generated Images}

The proliferation of AI-generated media has raised the question of whether humans are able to distinguish authentic visual content from synthetic imagery. Earlier work in this area, particularly prior to recent advances in large vision models, largely focused on people's ability to differentiate real human faces from those generated through synthetic methods. One such study found that participants often performed no better than chance and frequently mistook synthetic faces for real ones \citep{shen_study_2021}. Moreover, while earlier generations of synthetic faces were more readily detected, faces produced using newer techniques increasingly deceived observers and were sometimes judged as more realistic than actual photographs \citep{lago_more_2022}.

Subsequent work, particularly following the emergence of large vision models, has given rise to a growing body of research examining humans' ability to identify AI-generated images. Studies addressing this question across different data types report mixed results. For example, one study investigated participants' ability to distinguish AI-generated images from real photographs across everyday categories such as landscapes, architecture, and interiors. Although overall accuracy was moderate, detection rates declined substantially for images produced by more advanced models \citep{hogemann_what_2025}. Another study explored how users perceive, trust, and evaluate AI-generated content on social media. It reported cautious acceptance of AI-generated content alongside a pronounced gap between participants' perceived and actual ability to distinguish AI-generated from human-made material \citep{labajova_state_2023}. Additional work has examined the perception and evaluation of AI-generated artworks in comparison to human-made and co-designed pieces. The study found that participants were generally able to identify AI-generated artworks with moderate accuracy \citep{rueda-arango_exploring_2024}. Finally, a large cross-country study found that participants struggled to distinguish AI-generated from human-produced media across images, audio, and text, often performing at or below chance level. Participants consistently rated AI-generated media as more likely to be human-produced, underscoring the increasing realism and persuasive power of contemporary generative models \citep{frank_representative_2024}.

\subsection{AI-Based Detection Methods for AI-Generated Images}

Research on the use of AI models for AI-generated image detection is rapidly expanding.  \cite{mahara_methods_2026} reviewed recent methods for detecting AI-generated images as generative models become increasingly realistic and widespread. It categorized detection approaches into seven groups, ranging from spatial and frequency analysis to training-free and multimodal reasoning-based methods, and highlighted a growing shift toward multimodal and hybrid frameworks.  \cite{zhang_vision-language_2024} examined vision--language models (VLMs) as an emerging paradigm that leverages large-scale image--text data to enable zero-shot, open-vocabulary visual recognition, reducing reliance on task-specific labeled datasets and fine-tuning. It found that while vision--language models achieve strong zero-shot performance in image classification, they remain less effective in dense tasks such as detection and segmentation.

As suggested by these surveys, the literature on language-model-based image analysis is diverse. For instance, one study evaluated whether AI can reliably detect AI-generated images using a convolutional neural network framework to distinguish images generated by generative adversarial networks from real images, achieving near-perfect accuracy on benchmark datasets and strong generalization to unseen data \citep{baraheem_ai_2023}. Another proposed a zero-shot detection method that requires neither synthetic images nor knowledge of specific generators during training. By modeling the statistical distribution of real images with a lossless image encoder, the method flags images that significantly deviate from this distribution and demonstrates strong generalization to unseen generators \citep{leonardis_zero-shot_2025}. A related study investigated detection in an online setting where new generators emerge over time, showing that incremental training on previously released models improves performance on future unseen models, although major architectural shifts still degrade accuracy \citep{epstein_online_2023}. Another work explored the use of contrastive language--image pre-training for AI-generated image detection and showed that a lightweight, few-shot contrastive language--image pre-training--based detector generalizes well across a wide range of generators \citep{cozzolino_raising_2024}. Finally,  \cite{park_performance_2024} found that artifact-based approaches perform best on images generated by generative adversarial networks, while image-encoder-based methods generalize more effectively to diffusion- and transformer-based generators.

\section{Methodology}
\label{sec:method}

\begin{figure*}[!th]
    \centering
    \includegraphics[width=\textwidth]{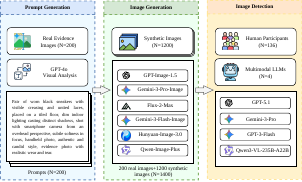}
    \caption{Overview of the experimental design. Real evidence images ($N=200$) are paired with AI-generated counterparts from six generation models ($N=1,200$), creating a dataset of 1,400 images. Human participants ($N=136$) and MLLMs ($N=4$) perform the same detection task, enabling direct comparison.}
    \label{fig:overview}
\end{figure*}

To address our research questions, we constructed a novel dataset, then use an experimental design consisting of two parallel studies. \textbf{Study~1} investigates human detection performance through a controlled web-based experiment, while \textbf{Study~2} evaluates MLLMs under comparable conditions. Both studies use identical stimuli and response formats, enabling direct comparison between human and AI detection capabilities. This section first describes dataset construction and then details the two studies. Figure~\ref{fig:overview} presents the overall experimental design. 

\subsection{Dataset Construction}

We refer to our dataset as SLED-1400. To our knowledge, SLED-1400 is the first object-centric dataset that pairs authentic visual evidence with AI-generated counterparts in legally relevant contexts. The dataset covers the categories of visual evidence most frequently submitted in everyday legal disputes. Unlike prior synthetic media datasets, such as FaceForensics++ \citep{rossler2019faceforensics++} and DFDC \citep{dolhansky2020deepfake}, which are primarily face-centric, SLED-1400 is tailored to legal applications. Each synthetic image is semantically paired with an authentic photograph sharing the same scene content and visual semantics, providing real-world grounding and enabling controlled comparisons between authentic and AI-generated evidence. In addition, SLED-1400 includes images generated by six contemporary text-to-image models, allowing for cross-generator analysis. The full dataset, prompt-generation outputs, model responses, and human-response logs will be released upon acceptance.

We first assembled a dataset of real-world evidentiary images reflecting the types of visual materials commonly submitted in civil disputes and then constructed a matched set of synthetic counterparts generated from these real images. We now describe how we collected authentic visual evidence and reproduced it using state-of-the-art generative models to construct these paired comparisons.

\subsubsection{Authentic Evidence Image Collection}

Evidence used in legal dispute resolution is inherently diverse, as it must support a wide range of legal arguments across different case types and fact patterns. However, specialized visual evidence datasets tailored to legal disputes remain rare. To address this gap, we identified ten categories of visual evidence frequently encountered in civil disputes, particularly in consumer disputes, insurance claims, and product liability cases. These categories are highly representative of ``high-volume, low-value'' disputes, where automated or AI-assisted evidence verification is especially promising for improving efficiency. Table~\ref{tab:image_categories} presents our complete category taxonomy.

Following this taxonomy, we manually selected authentic images from multiple public repositories (e.g., datasets hosted on Hugging Face) and targeted web searches to ensure high domain relevance. We collected 20 images per category, resulting in a core dataset of 200 authentic evidentiary images. The screening process prioritized authenticity and evidentiary realism. We explicitly excluded professionally staged photographs with idealized lighting/exposure, pronounced bokeh (e.g., shallow depth of field), or studio-quality product shots. Instead, we selected images reflecting the characteristics of evidence submitted in disputes and legal proceedings, including varied lighting conditions, different capture devices and camera quality (e.g., from recent high-end smartphones to older devices), and natural compositional flaws (e.g., motion blur or poor framing).

\begin{table}[t]
    \centering
    \begin{tabular}{@{}lp{8cm}@{}}
        \toprule
        \textbf{Category} & \textbf{Description} \\
        \midrule
        Product Defects & Manufacturing defects, product failures, quality issues \\
        Food Spoilage & Contaminated food, foreign matter in products \\
        Clothing Defects & Fabric damage, stitching flaws, material defects \\
        Electronics Damage & Device failures, screen damage, electrical issues \\
        Household Damage & Furniture damage, household product failures \\
        Car Damage & Vehicle damage, scratches, dents, collision evidence \\
        Receipts \& Lists & Documentary evidence, written records, invoices \\
        Delivery Scene & Delivery documentation, package placement photos \\
        Environmental Damage & Weather-related damage, environmental incidents \\
        Plant Damage & Plant disease, pest damage, agricultural evidence \\
        \bottomrule
    \end{tabular}
    \caption{Evidence categories used in the dataset}
    \label{tab:image_categories}
\end{table}

\subsubsection{Synthetic Evidence Image Generation}

In addition to the authentic evidence image collection, we generated a synthetic image dataset paired with the authentic image dataset using a standardized, reproducible pipeline. The pipeline involves two rounds of Application Programming Interface (API) calls. First, we use GPT-4o \citep{hurst2024gpt} to analyze and describe each authentic image and to produce a reconstruction-oriented prompt intended to reproduce the key visual content of the image. The process produces 200 prompts in total (one prompt per authentic image). Second, we feed these prompts into six state-of-the-art text-to-image models selected from the Text-to-Image leaderboard LM Arena \citep{chiang2024chatbot} to generate corresponding synthetic images, producing 1,200 AI-generated images in total (200 prompts $\times$ 6 models). We provide a more detailed description of the generation pipeline below.

\textbf{Stage 1: Prompt Generation.}
We provided the prepared \textit{Prompt Generation Requirement} together with each authentic image to GPT-4o, instructing the model to analyze the image and generate a structured description suitable for recreating a visually and semantically similar evidentiary photograph. The model was explicitly guided to describe the image from the perspective of evidence documentation rather than artistic rendering, capturing both factual content and realistic visual style.

This structured representation serves two purposes. First, it enables the preservation of evidentiary realism by explicitly encoding common imperfections present in real-world evidence images (e.g., suboptimal lighting, limited camera quality, or incidental background clutter). Second, it constrains downstream text-to-image models to avoid non-evidentiary or stylized outputs (such as cartoons, posters, or graphic illustrations), thereby keeping the generated images aligned with the conventions of authentic evidence photography.

For each authentic image, GPT-4o produced a structured output containing the following fields:
\begin{itemize}
\item \textbf{Category}: the assigned evidence category based on our taxonomy.
\item \textbf{Description}: a concise natural-language summary of the overall scene and evidentiary context.
\item \textbf{KeyElements}: a list of salient evidential elements critical for dispute resolution (e.g., damage patterns or defect characteristics).
\item \textbf{VisualDetails}: a structured specification of visual attributes, including colors, lighting conditions, background context, main subject, camera perspective, and observable imperfections.
\item \textbf{ImageGenerationPrompt}: a consolidated prompt synthesized from the above fields, explicitly designed for text-to-image models to generate a realistic, documentary-style evidentiary image.
\end{itemize}

The resulting \texttt{\allowbreak ImageGeneration\allowbreak Prompt} integrates semantic content, evidential salience, and visual constraints into a single instruction, ensuring that the generated images faithfully reflect both the factual elements and the non-ideal visual qualities characteristic of real-world evidence submissions.

\textbf{Stage 2: Image Synthesis.}
After obtaining the \texttt{\allowbreak ImageGeneration\allowbreak Prompt} from Stage~1, we used this prompt to generate synthetic evidence images with six state-of-the-art text-to-image models. The same base prompt was provided to all models, though minor model-specific 
adaptations were necessary: Hunyuan truncated prompts exceeding 1,024 characters. No negative prompts were used.

For image resolution, four generation models (GPT-Image-1.5, Flux-2-Max, Hunyuan-3.0, and Qwen-Image-Plus) were configured to output 1024$\times$1024 images, while the two Gemini models used their default resolution settings. Additionally, Qwen-Image-Plus was set to photographic style mode, and Hunyuan enabled automatic prompt refinement.

For each prompt, each model generated a single image, resulting in one synthetic image per prompt-model pair. No manual selection or curation was performed. In cases where a generation attempt failed (e.g., incomplete output or system errors), the image was regenerated using the same prompt until a valid image was produced.

No post-processing was applied to the generated images, with one exception. The Hunyuan-Image-3.0 model embeds a visible watermark in the lower-right corner of generated images. To ensure consistency across models and avoid introducing model-specific visual artifacts, we uniformly masked all images generated by the Hunyuan-Image-3.0 model, removing the bottom 8\% of the image height.

Six generation models were employed to create synthetic evidence images:

\begin{itemize}
    \item \textbf{GPT-Image-1.5} (OpenAI): OpenAI's latest image generation model (t the time of data collection) with enhanced photorealism capabilities
    \item \textbf{Gemini-3-Pro-Image} (Google): Google's flagship multimodal generation model
    \item \textbf{Flux-2-Max} (Black Forest Labs): A leading open-weight diffusion model known for high-fidelity outputs
    \item \textbf{Gemini-2.5-Flash-Image} (Google): Google's efficient generation model optimized for speed
    \item \textbf{Hunyuan-Image-3.0} (Tencent): Tencent's advanced image synthesis model
    \item \textbf{Qwen-Image-Plus} (Alibaba): Alibaba's multimodal generation model
\end{itemize}

The selected models collectively span open-source and closed-source systems, represent both Chinese and U.S. model ecosystems, and cover a range of design objectives, including high-fidelity generation versus efficiency-oriented generation, as well as different training paradigms. This diversity allows for a systematic comparison of synthetic evidence images produced by contemporary text-to-image models under a unified prompting and evaluation framework.

\subsection{Study 1: Human Detection Experiment}

Study 1 measures human detection performance in a controlled web-based experiment. In practical contexts, the primary encounters with digital evidence involve laypeople, such as consumers, merchants, platform moderators, and litigants in high-volume, low-value disputes. These individuals represent a demographic that routinely evaluates visual evidence without the benefit of formal forensic expertise. This is also the demographic for which AI-generated synthetic media is most consequential. Research on memory distortion suggests that fabricated evidence can induce false memories or precipitate wrongful accusations by rendering fictitious events \citep{iannuzzi_ai-generated_2026}.


Therefore, to accurately measure the detection capabilities of this common demographic, we intentionally recruited participants without specialized forensic training. We imposed no restrictions on professional or educational backgrounds, as our objective was to determine whether individuals from diverse everyday contexts without specialized training can reliably distinguish authentic evidentiary photographs from AI-generated fabrications.

\paragraph{Participants.} We recruited \textbf{136 adult participants} online, who voluntarily completed the experiment. Before beginning, all participants were informed that the study pertained the authenticity of evidentiary images and that a subset of these images might be AI-generated, and provided informed consent. Of 205 individuals who entered the standard arm of the experiment, 136 completed all 20 trials and form the analytical sample ($N = 2{,}720$ trial responses). The 69 participants who began but did not complete the session are excluded from all reported statistics.

\paragraph{Apparatus.} To ensure consistency across experimental conditions, we developed a web-based experimental platform called SLED, built using Next.js and TypeScript. Images were displayed within a responsive container with constrained dimensions (280--450px height) while preserving original aspect ratios, zooming and other image manipulations were disabled. This setup serves two purposes. First, it maintains experimental control by ensuring participants view stimuli under near-identical conditions regardless of device. Second, it enhances ecological validity. By mirroring the constrained, non-interactive viewing environments typical of online dispute resolution platforms, e-commerce systems, and insurance portals, this methodological choice aims to reflect how visual evidence is actually assessed in high-volume professional settings.

\paragraph{Procedure and sampling.} Each participant completed a single session of 20 trials: 10 authentic images and 10 synthetic images. To prevent any single evidence category from dominating a participant's session while still preserving randomness, we applied a soft category-diversity sampling procedure. The platform first sampled candidate image families to encourage broad category coverage within the session, and then assigned 10 trials each to the authentic and synthetic conditions. This procedure promotes within-session category diversity but does not enforce strict category matching between the authentic and synthetic trials a participant sees. For synthetic trials, the sampler prioritized generative models that the participant had not yet encountered in the session, increasing per-participant coverage of diverse generation methods. When all six models had already been seen, the platform fell back to uniform random draw among available synthetic variants from the sampled family.

\paragraph{Response schema.} Participants viewed one stimulus at a time and provided multidimensional responses: (1) an authenticity judgement (\textit{Real}, \textit{Synthetic}, or \textit{Unsure}); (2) a confidence rating on a 7-point Likert scale (1 = guessing, 7 = completely certain); and (3) one to three reason tags drawn from the predefined vocabulary in Table~\ref{tab:reasons}. A free-text field was available for optional additional explanation. The \textit{Unsure} option was included to minimize forced guessing, which would otherwise inject noise into accuracy estimates. Of the 2{,}720 trial responses recorded, 146 (5.4\%) were \textit{Unsure}, these responses are excluded from accuracy and signal-detection statistics but are reported separately as a metric of expressed uncertainty.

Throughout the paper, we map the 7-point confidence rating to a predicted probability of correctness via
\[
p_{\text{pred}} \;=\; 0.5 + \frac{c - 1}{12}, \qquad c \in \{1,\dots,7\},
\]
so that $c = 1$ corresponds to $p_{\text{pred}} = 0.5$ (pure guess) and $c = 7$ corresponds to $p_{\text{pred}} = 1.0$ (complete certainty). This mapping is used in all calibration analyzes (Section~\ref{sec:results}), including the reliability diagrams and the Expected Calibration Error and Brier-score computations.

\begin{table}[t]
    \centering
    \begin{tabular}{@{}cl@{}}
        \toprule
        \textbf{ID} & \textbf{Reason Tag} \\
        \midrule
        1 & Unnatural texture or material \\
        2 & Inconsistent lighting or shadows \\
        3 & Abnormal edges or contours \\
        4 & Incorrect structural details \\
        5 & Abnormal text or symbols \\
        6 & Unreasonable perspective or spatial relations \\
        7 & Background repetition or generation artifacts \\
        8 & Content contradicts common sense \\
        9 & Looks like a real photo \\
        10 & Intuitive, hard to explain \\
        \bottomrule
    \end{tabular}
    \caption{Reason tags for authenticity judgments}
    \label{tab:reasons}
\end{table}

\subsection{Study 2: MLLM Detection Experiment}

In Study 2, we evaluated four state-of-the-art MLLMs using the same SLED-1400 stimulus, using the same response schema as in Study 1 (choice, confidence, 1--3 reason tags, and free-text reasoning). We selected top-ranked models (during the time of the experiment) from the Vision leaderboard of LM Arena \citep{chiang2024chatbot}:

\begin{enumerate}
    \item \textbf{GPT-5.1} (OpenAI): OpenAI's latest multimodal model with advanced visual reasoning
    \item \textbf{Gemini 3 Pro} (Google): Google's flagship multimodal model
    \item \textbf{Gemini 3 Flash} (Google): Google's efficient multimodal model optimized for throughput
    \item \textbf{Qwen3 VL 235B A22B Instruct} (Alibaba): Alibaba's large-scale multimodal model with 235 billion parameters
\end{enumerate}

All four models received the same system prompt (see Prompt~\ref{prompt:vlm-system}), the prompt included the reason-tag vocabulary shown to human participants (Table~\ref{tab:reasons}), and were queried once per image. Inference parameters were standardized across model, with \texttt{temperature = 0} for near-deterministic outputs and \texttt{max\_tokens = 4096}. Each model returned exactly one response per image (no resampling, no chain-of-thought retries), producing 5{,}600 MLLM responses ($4$ models $\times$ $1{,}400$ images) in total. All queries were routed through a unified API gateway (OpenRouter). Unlike Study 1, in which each participant saw 20 stimuli, each MLLM evaluated every image in SLED-1400.

\refstepcounter{prompt}
\begin{promptbox}
\label{prompt:vlm-system}
\begin{promptverbatim}
You are an expert at analyzing images to determine if they are real photographs or AI-generated.

Your task is to examine the given image and determine:
1. Whether it is a REAL photograph or AI-generated (SYNTHETIC), or if you are UNSURE
2. Your confidence level (1-7, where 1=very uncertain, 7=very certain)
3. The reasons for your judgment, selected from the predefined tags below
4. A brief explanation of your reasoning

Available reason tags:
[The predefined reason-tag list was inserted here.]

You must respond in valid JSON format with this exact structure:
{
  "choice": "REAL" | "SYNTHETIC" | "UNSURE",
  "confidence": 1-7,
  "reasonTagIds": [array of 1-3 tag IDs from the list above],
  "reasoning": "Brief explanation of your judgment (1-2 sentences)"
}
\end{promptverbatim}
\end{promptbox}

For a complementary check against tools built specifically for AI-image detection, we evaluated three detectors (UnivFD \citep{ojha2023fakedetect}, CNNDetect \citep{wang2020cnn}, and a public fine-tuned ViT detector \citep{li2022exploring}) on SLED-1400 dataset. Each detector outputs a probability score for SYNTHETIC. We apply a 0.5 threshold and compute the same accuracy and balanced-accuracy metrics used for the human and MLLM evaluators. Results are reported in Section~\ref{sec:specialized-detectors}.

\subsection{Free-text Reasons Coding}
\label{sec:reason-coding}

To complement the categorical-judgement analysis in Section~\ref{sec:results}, we coded the free-text reasons that raters provided alongside each judgement. Across the 2{,}720 trial responses in Study~1, human participants wrote at least one free-text reasoning in 1{,}574 trials (57.9\,\%), and all responses were translated into English for analysis. In Study~2, every MLLM response included a brief reason paragraph, resulting in a total of 5{,}599 reasons (only one of the 5{,}600 responses returned an empty string). 

We coded the free-text reasons using a mixed deductive and inductive approach \citep{braun2006using}. The initial codebook used the ten reason categories in Table~\ref{tab:reasons} as root codes. We then added three codes identified from a pilot sample of 200 reasons: \emph{no-anomaly default template}, for responses that stated that no AI artifacts were visible without giving a concrete image-based observation; \emph{domain expertise}, for appeals to personal or professional knowledge; and \emph{explicit hedging}, for explicit expressions of uncertainty. Each reason received one to three primary codes. We also recorded whether it mentioned an image-specific detail, meaning a detail that could be checked directly against the image, and whether its cues pointed toward a \textit{real}, \textit{synthetic}, \textit{mixed}, or \textit{none} judgement. Coding was performed with Claude Sonnet 4.6 (to avoid confounding effects from models within the same family) using \texttt{temperature = 0} and a fixed codebook schema. 

\section{Results}
\label{sec:results}

This section presents the empirical findings. The human sample consisted of 136 participants, who produced 2{,}720 responses in total. These responses include both decided judgements and \textit{Unsure} responses. Among the 2{,}720 human responses, 146 were \textit{Unsure}, corresponding to an unsure rate of 5.4\%. The MLLM evaluation produced 5{,}600 model responses in total. Only 9 of these responses were \textit{Unsure}, and all 9 came from GPT-5.1. In the accuracy analyses, percentages are calculated on the decided subset of trials. We also report \textit{balanced accuracy} alongside raw overall accuracy. Unsure rates are reported separately.

This section is structured as follows. Section~\ref{sec:human-modes} asks whether and how lay humans fail at detecting AI-generated visual evidence. Section~\ref{sec:mllm-modes} asks the same question of frontier MLLMs. Section~\ref{sec:specialized-detectors} reports the performance of detectors built specifically for AI-image classification. Section~\ref{sec:combination} asks whether the two failure modes are independent enough that a combined workflow could outperform either alone. Section~\ref{sec:reason-patterns} analyzes the reasons raters provide for their decisions from a qualitative perspective. Table~\ref{tab:performance-overview} gives the headline numbers.

\begin{table*}[!htbp]
\centering
\small
\renewcommand{\arraystretch}{1.15}
\resizebox{\textwidth}{!}{%
\begin{tabular}{ll cc cc cccccc}
\toprule
\multirow{2}{*}{\textbf{Category}} & \multirow{2}{*}{\textbf{Evaluator}} & \textbf{Ovr.} & \textbf{Bal.} & \multicolumn{2}{c}{\textbf{Base Metrics}} & \multicolumn{6}{c}{\textbf{Synthetic Image Detection Rate by Generator}} \\
\cmidrule(lr){5-6} \cmidrule(lr){7-12}
 & & \textbf{Acc.} & \textbf{Acc.} & Real & Synth. & Flux-2 & Gem-2.5 & Gem-3 & GPT-1.5 & Huny-3 & Qwen \\
\midrule
\multicolumn{2}{l}{\textbf{Human}} &
\cellcolor{hcolor!65}\textbf{0.648} & \cellcolor{hcolor!65}\textbf{0.648} & \cellcolor{hcolor!64}\textbf{0.642} & \cellcolor{hcolor!65}\textbf{0.654} & \cellcolor{hcolor!51}\textbf{0.510} & \cellcolor{hcolor!62}\textbf{0.624} & \cellcolor{hcolor!49}\textbf{0.485} & \cellcolor{hcolor!53}\textbf{0.525} & \cellcolor{hcolor!83}\textbf{0.831} & \cellcolor{hcolor!93}\textbf{0.930} \\
\midrule
\multirow{4}{*}{\rotatebox{90}{MLLMs}}
 & Gemini-3-Flash &
 \cellcolor{hcolor!64}0.638 & \cellcolor{hcolor!79}0.789 & \cellcolor{hcolor!100}1.000 & \cellcolor{hcolor!58}0.578 & \cellcolor{hcolor!35}0.350 & \cellcolor{hcolor!70}0.700 & \cellcolor{hcolor!14}0.140 & \cellcolor{hcolor!28}0.280 & \cellcolor{hcolor!100}0.995 & \cellcolor{hcolor!100}1.000 \\
 & Gemini-3-Pro &
 \cellcolor{hcolor!60}0.604 & \cellcolor{hcolor!77}0.769 & \cellcolor{hcolor!100}1.000 & \cellcolor{hcolor!54}0.538 & \cellcolor{hcolor!35}0.350 & \cellcolor{hcolor!58}0.580 & \cellcolor{hcolor!7}0.070 & \cellcolor{hcolor!25}0.245 & \cellcolor{hcolor!99}0.985 & \cellcolor{hcolor!100}1.000 \\
 & GPT-5.1 &
 \cellcolor{hcolor!42}0.419 & \cellcolor{hcolor!66}0.661 & \cellcolor{hcolor!100}1.000 & \cellcolor{hcolor!32}0.322 & \cellcolor{hcolor!4}0.035 & \cellcolor{hcolor!18}0.175 & \cellcolor{hcolor!2}0.015 & \cellcolor{hcolor!5}0.050 & \cellcolor{hcolor!72}0.718 & \cellcolor{hcolor!95}0.954 \\
 & Qwen3-VL-235B &
 \cellcolor{hcolor!42}0.421 & \cellcolor{hcolor!66}0.662 & \cellcolor{hcolor!100}1.000 & \cellcolor{hcolor!32}0.324 & \cellcolor{hcolor!6}0.055 & \cellcolor{hcolor!16}0.155 & \cellcolor{hcolor!1}0.010 & \cellcolor{hcolor!2}0.015 & \cellcolor{hcolor!76}0.760 & \cellcolor{hcolor!95}0.950 \\
\cmidrule{2-12}
 & \textbf{MLLM Average} &
\cellcolor{hcolor!52}\textbf{0.521} & \cellcolor{hcolor!72}\textbf{0.720} & \cellcolor{hcolor!100}\textbf{1.000} & \cellcolor{hcolor!44}\textbf{0.441} & \cellcolor{hcolor!20}\textbf{0.198} & \cellcolor{hcolor!40}\textbf{0.403} & \cellcolor{hcolor!6}\textbf{0.059} & \cellcolor{hcolor!15}\textbf{0.148} & \cellcolor{hcolor!86}\textbf{0.864} & \cellcolor{hcolor!98}\textbf{0.976} \\
\bottomrule
\end{tabular}%
}
\caption{Performance comparison of humans (\textit{N}~=~136 participants, 2{,}574 decided trials) and MLLMs (5{,}591 decided responses) on SLED-1400. ``Ovr.\ Acc.'' is sample-weighted; ``Bal.\ Acc.'' is the arithmetic mean of real- and synthetic-image accuracy.}
\label{tab:performance-overview}
\end{table*}

\subsection{How Lay Humans Fail}
\label{sec:human-modes}

Among all the decided judgements (2{,}574), human participants' overall performance is 64.8\%, above the chance baseline, but far from the reliability required for legal and dispute-related decisions. Humans miss synthetic content 34.6\,\% of the time and wrongly reject real images 35.8\,\% of the time. Sensitivity (65.5\,\%) and specificity (64.2\,\%) are roughly equal, which means that adjusting a single decision threshold will not improve overall accuracy. Humans are neither overly conservative nor overly aggressive, they simply make errors in both directions. 

\subsubsection{Which categories and generators make humans fail}
\label{sec:human-axes}

Our results suggest the factors that may influence human performance include the category of evidence and the generator that produced the synthetic image.

Table~\ref{tab:f12-fp} reports both human error rates --- wrongly rejecting a real image (FA rate) and wrongly accepting a synthetic image (Miss rate) --- for each of the ten evidence categories. Every category exceeds a 27.5\,\% FA rate, and there is no category in the corpus on which lay humans reliably accept real images.

The categories with the highest FA rate are typically not the categories with the highest Miss rate. Across the ten categories, the two rates run in opposite directions (Pearson $r = -0.60$). Receipts \&\ lists are the worst case for over rejection. 51.3\,\% of real receipts were incorrectly classified as synthetic ones, but on synthetic receipts, humans missed only 29.1\,\%, the lowest miss rate in the corpus. Electronics damage shows opposite patterns. Humans rarely over-rejected real electronic damage images (31.2\,\% FA), but missed 43.1\,\% of the synthetic ones. 

The category-level pattern therefore does not reflect uniform noise in both directions. Instead, human errors were category-specific. Some categories, such as receipts, pushed participants toward over-suspicion, whereas others, such as electronics damage, pushed them toward over-trust. These two biases have different implications for evidence law. Over-suspicion risks discrediting real evidence, while over-trust risks admitting fake evidence.

\begin{table}[!htbp]
  \centering
  \small
  \setlength{\tabcolsep}{6pt}
  \renewcommand{\arraystretch}{1.15}
  \begin{tabular}{lccc}
    \toprule
    \textbf{Evidence category} & \shortstack{\textbf{Sample size}\\\textbf{(real / synth.)}} & \shortstack{\textbf{FA rate}\\\textbf{[95\,\% CI]}} & \shortstack{\textbf{Miss rate}\\\textbf{[95\,\% CI]}} \\
    \midrule
    Receipts \& lists                & 119 / 141 & \textbf{51.3\,\%} [42.4, 60.1] & 29.1\,\% [22.2, 37.0] \\
    Environmental damage  & 197 / 192 & 38.6\,\% [32.1, 45.5] & 27.6\,\% [21.8, 34.3] \\
    Car damage                       & 142 / 133 & 38.0\,\% [30.5, 46.2] & 27.8\,\% [20.9, 36.0] \\
    Food spoilage                    & 135 / 177 & 37.0\,\% [29.4, 45.4] & 39.0\,\% [32.1, 46.3] \\
    Plant damage                     & 143 / 123 & 35.7\,\% [28.3, 43.8] & 37.4\,\% [29.4, 46.2] \\
    Clothing defects                 & 160 / 154 & 32.5\,\% [25.7, 40.1] & 38.3\,\% [31.0, 46.2] \\
    Electronics damage               & 128 / 109 & 31.2\,\% [23.9, 39.7] & \textbf{43.1\,\%} [34.2, 52.5] \\
    Delivery scenes                  & 123 / 122 & 29.3\,\% [22.0, 37.8] & 36.9\,\% [28.8, 45.7] \\
    Household damage                 &  28 /  43 & 28.6\,\% [15.3, 47.1] & 37.2\,\% [24.4, 52.1] \\
    Product defects                  &  91 / 114 & 27.5\,\% [19.4, 37.4] & 35.1\,\% [26.9, 44.2] \\
    \midrule
    \textbf{Overall}                 & \textbf{1{,}266 / 1{,}308} & \textbf{35.8\,\%} [33.2, 38.5] & \textbf{34.6\,\%} [32.0, 37.3] \\
    \bottomrule
  \end{tabular}
  \caption{Human error rates on real and synthetic images, by evidence category (decided trials only). Categories sorted by FA rate (descending).}
  \label{tab:f12-fp}
\end{table}

\begin{figure}[!htbp]
    \centering
    \includegraphics[width=\columnwidth]{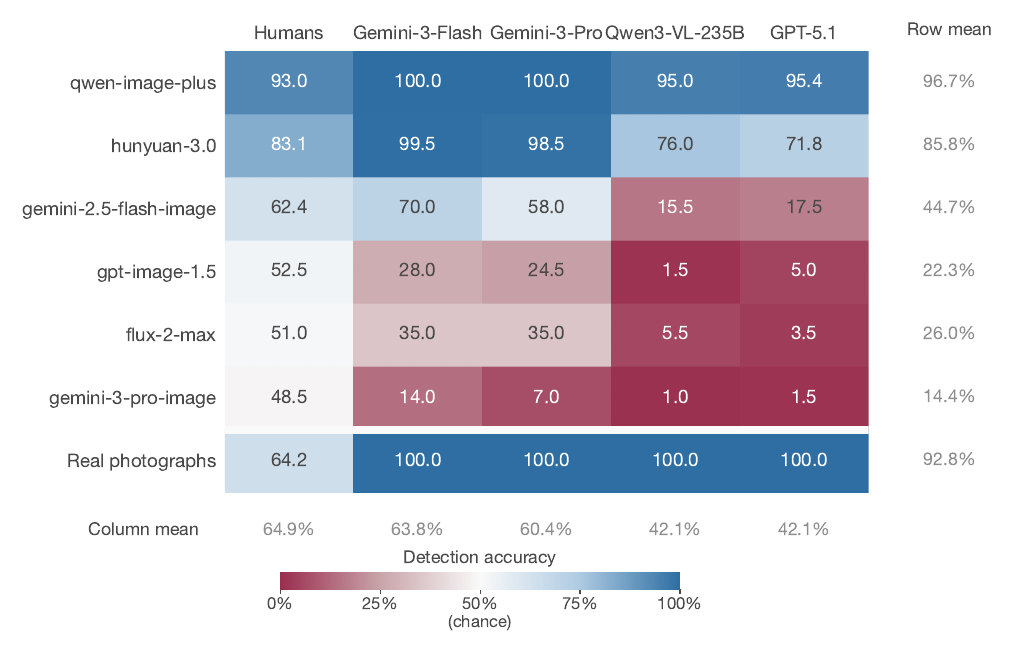}
    \caption{Synthetic-detection rate by generator (rows) and rater (columns).}
    \label{fig:f3-heatmap}
\end{figure}

Detectability varied dramatically across image generation models. As Figure~\ref{fig:f3-heatmap}, human detection rates span a range of 44.5 percentage points, from near-certain to near-random identification. This indicates that we are currently experiencing a turning point in the evolution of model capabilities.

Humans flagged Qwen-Image-Plus and Hunyuan-Image-3.0 outputs at 93.0\,\% and 83.1\,\%, but performance fell close to chance on the four harder generators: 48.5\,\% on Gemini-3-Pro, 51.0\,\% on Flux-2-Max, 52.5\,\% on GPT-Image-1.5, and 62.4\,\% on Gemini-2.5-Flash.

\subsubsection{What humans get right}
\label{sec:human-calibrated}

\begin{figure}[!htbp]
  \centering
  \includegraphics[width=\linewidth]{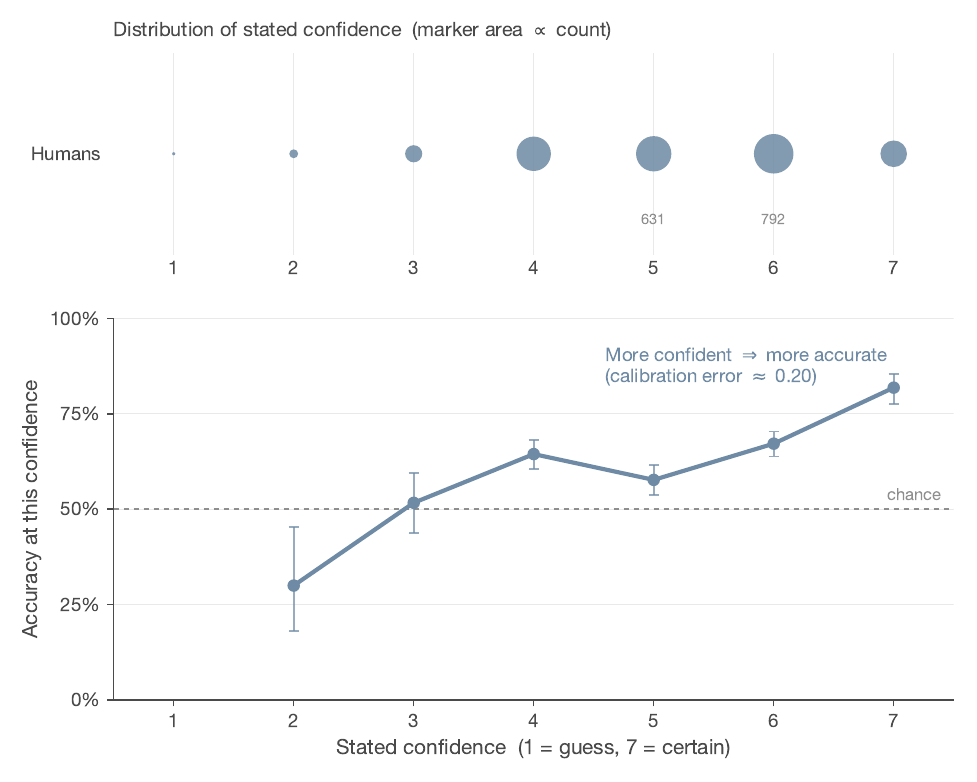}
  \caption{Human accuracy at each confidence level (1--7), with 95\,\% Wilson CIs (bottom panel); distribution of stated confidence (top panel, mean 5.18).}
  \label{fig:f13a}
\end{figure}

Humans' stated confidence tracks how often they are correct. Figure~\ref{fig:f13a} shows accuracy at each confidence level: it rises from about 30\,\% at confidence 2 to 82\,\% at confidence 7. The calibration error is 0.20 (lower being better), which is the average gap between stated confidence and actual accuracy. As Section~\ref{sec:mllm-modes} shows, that number is well below every MLLM we evaluated.

Humans use the \textit{Unsure} option in a useful direction. The overall abstention rate is 5.4\,\%, and 64.4\,\% of those abstentions (94 of 146) landed on real images. When lay humans cannot decide, they hedge toward not rejecting a real image. When hedging, the MLLMs shows the opposite behavior. Three of the four models never choose \textit{Unsure}, and GPT-5.1 abstained on only 9 of 1{,}400 trials (0.6\,\%), all on synthetic images (see Section~\ref{sec:mllm-calibration}).

Two further patterns in human responses support the same picture. Humans did not improve within a 20-trial session. Accuracy on the first five trials (63.2\,\%) is statistically indistinguishable from the last five (64.2\,\%). Interestingly, response time (RT) also runs against accuracy (see Table~\ref{tab:f14-snap}). Accuracy actually declines from 72.0\,\% in the fastest quintile (median 12\,s) to 60.6\,\% in the slowest (median 123\,s). A long response time signals an item the rater cannot resolve, not one that benefits from more thought.

\begin{table}[!htbp]
  \centering
  \small
  \setlength{\tabcolsep}{8pt}
  \renewcommand{\arraystretch}{1.15}
  \begin{tabular}{lrrc}
    \toprule
    \textbf{Response-time quintile} & \textbf{$n$} & \textbf{Median RT (s)} & \textbf{Accuracy [95\,\% CI]} \\
    \midrule
    Q1 (fastest) & 515 &  12.3 & 72.0\,\% [68.0, 75.7] \\
    Q2           & 515 &  25.1 & 66.6\,\% [62.4, 70.5] \\
    Q3           & 514 &  42.2 & 64.0\,\% [59.8, 68.0] \\
    Q4           & 515 &  68.0 & 60.8\,\% [56.5, 64.9] \\
    Q5 (slowest) & 515 & 123.3 & 60.6\,\% [56.3, 64.7] \\
    \bottomrule
  \end{tabular}
  \caption{Human accuracy by response-time quintile (decided trials only).}
  \label{tab:f14-snap}
\end{table}

\paragraph{Key takeaways.} Lay humans make errors in both directions and cannot reliably distinguish real photographs from AI-generated ones on their own. Their confidence tracks their accuracy, their failures cluster on specific categories and specific generators in predictable ways, and when they are unsure they hedge toward not rejecting a real photograph. The shape of this failure --- two-sided errors, well-calibrated confidence, an abstention pattern biased toward accepting real images --- preserves several procedural properties that evidence law assumes.

\subsection{How Frontier MLLMs Fail}
\label{sec:mllm-modes}

Each of the four MLLMs rated all 1{,}400 SLED-1400 images, producing 5{,}591 decided responses and 9 abstentions, all from GPT-5.1. Overall accuracy ranged from 41.9\,\% for GPT-5.1 to 63.8\,\% for Gemini-3-Flash. The Gemini models were roughly on par with humans in raw accuracy, whereas GPT-5.1 and Qwen3-VL-235B performed notably worse. However, these overall accuracy figures mask a more important failure pattern, which will be described in the following three subsections.

\subsubsection{MLLMs over-trust synthetic images}
\label{sec:mllm-overall}

\begin{figure}[!htbp]
  \centering
  \includegraphics[width=\linewidth]{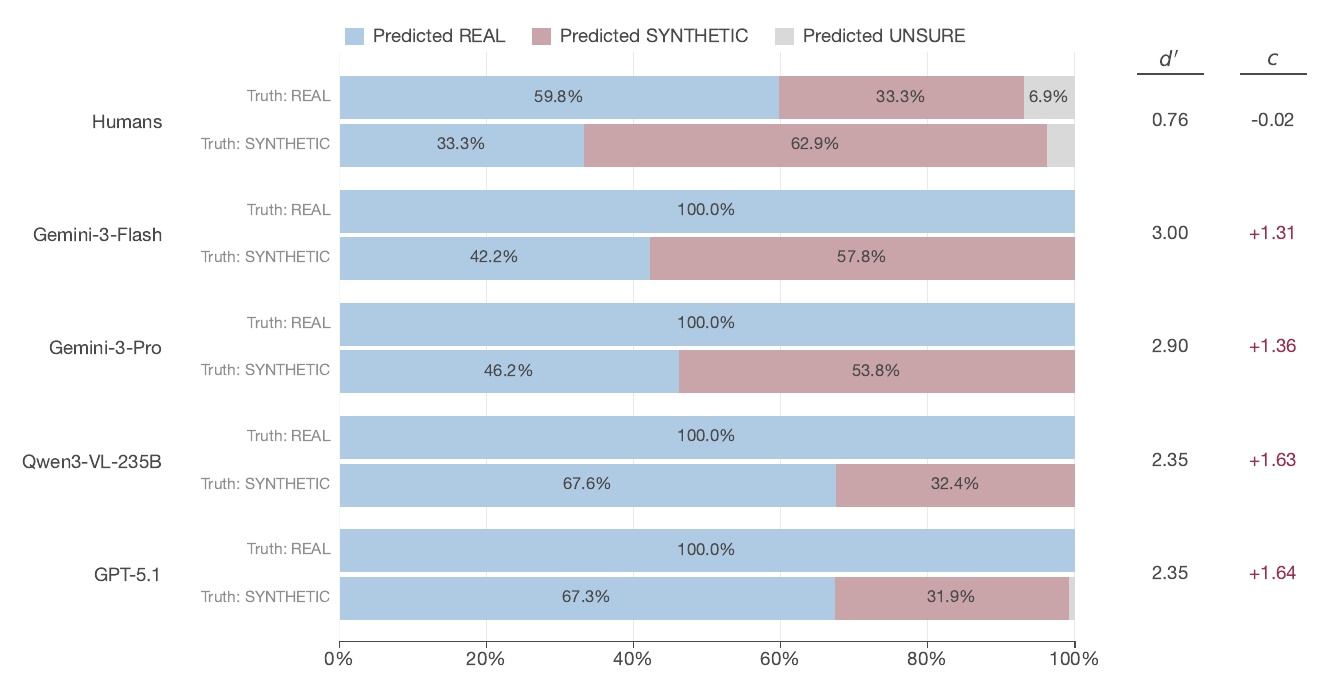}
  \caption{For each rater, the \textit{Truth: REAL} bar shows responses on real images and the \textit{Truth: SYNTHETIC} bar shows responses on synthetic images, broken into Predicted REAL (blue), Predicted SYNTHETIC (red), and Predicted UNSURE (grey). Right-hand columns: sensitivity $d'$ (higher = better separation between real and synthetic) and decision criterion $c$ (positive = bias toward calling an image real).}
  \label{fig:f2-bias}
\end{figure}

Surprisingly, across $4 \times 200 =800$ decisions on real images, not a single MLLM model misclassified a real image as a synthetic one. All four models achieved a specificity of 100\,\%. Sensitivity on synthetic images is 32\,\% (GPT-5.1) up to 58\,\% (Gemini-3-Flash) (see Figure~\ref{fig:f2-bias}). This is the opposite pattern from humans.

This pattern helps explain why some model accuracies fall far below 50\,\%. In a balanced real-versus-synthetic task, random guessing would be expected to produce accuracy around 50\,\%. Accuracy as low as 31.9\,\% is therefore not merely a sign of random error or weak visual perception. It reflects a strongly cautious decision rule. The models require unusually strong evidence before calling an image synthetic, which prevents false rejection of real images but causes many synthetic images to be accepted as real. 

\subsubsection{Stronger generators expose MLLM failures}
\label{sec:mllm-generator}

Figure~\ref{fig:f3-heatmap} shows MLLM sensitivity by generator. On Gemini-3-Pro-Image outputs, MLLM sensitivity is 7.0\,\% (Gemini-3-Pro), 14.0\,\% (Gemini-3-Flash), 1.5\,\% (GPT-5.1) and 1.0\,\% (Qwen3-VL-235B), for a four-MLLM average of 5.9\,\%. On Flux-2-Max and GPT-Image-1.5 the averages are 19.8\,\% and 14.8\,\%. Only the two easier generators (Qwen-Image-Plus, Hunyuan-Image-3.0) push MLLM detection above 50\,\%. 

\begin{figure}[!htbp]
  \centering
  \includegraphics[width=\linewidth]{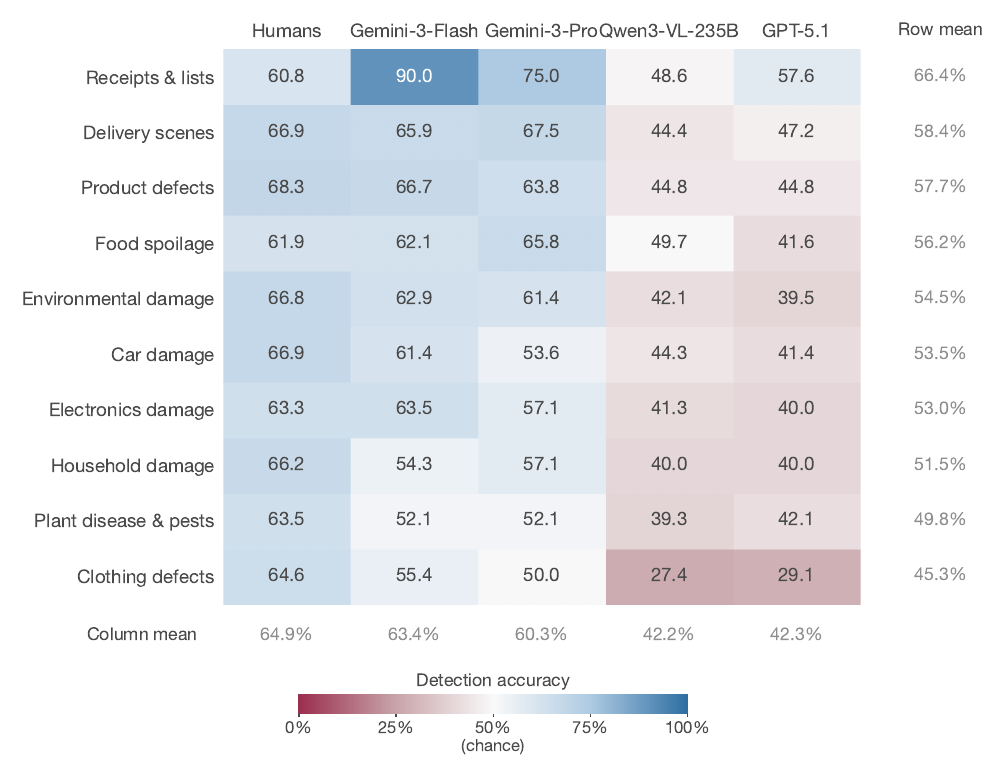}
  \caption{Per-rater accuracy on each of the ten evidence categories (real and synthetic trials combined, decided trials only). Rows sorted by row-mean accuracy across raters (descending); column means appear below each rater, row means in the right margin. Colour diverges around 50\,\% (chance).}
  \label{fig:f7-category}
\end{figure}

Figure~\ref{fig:f7-category} shows per-rater accuracy on each of the ten evidence categories, combining real and synthetic trials. Overall, MLLM performance is fairly consistent across categories, and varies much less than it does across different generators. In fact, for any given MLLM, the variation across categories is about half of the variation across generators shown in Figure~\ref{fig:f3-heatmap}. This suggests that the choice of generator matters more for an adversary than the type of image being used. 

A second finding is that humans and MLLMs struggle with different categories. The category that humans handle worst (Receipts \& Lists, 60.8\,\% accuracy) is the one where best MLLM (Gemini-3-Flash) performs best (90.0\,\%). On the other hand, Clothing defects is difficult for several MLLMs (Gemini-3-Flash, Gemini-3-Pro, and GPT-5.1), while humans perform close to their average on it. This difference in how categories affect performance adds another layer of divergence between human and MLLM error patterns, which we discuss further in Section~\ref{sec:combination}.

\subsubsection{Misplaced MLLM confidence}
\label{sec:mllm-calibration}

\begin{figure}[!htbp]
  \centering
  \includegraphics[width=\linewidth]{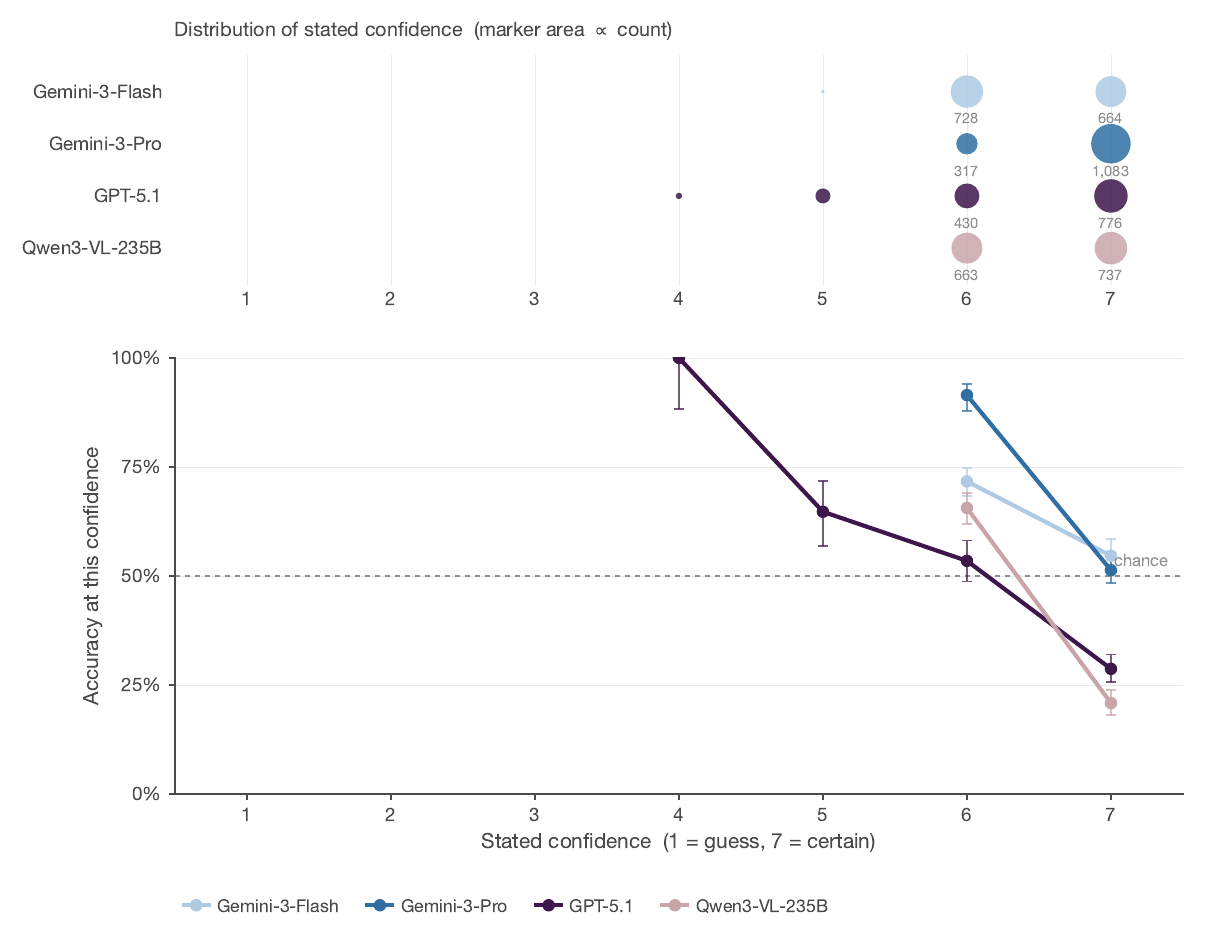}
  \caption{Top: distribution of stated confidence for each MLLM (marker area scales with response count). Bottom: accuracy at each confidence level. Dashed line: 50\,\% (chance).}
  \label{fig:f13b}
\end{figure}

The accuracy gap could, in principle, be reduced by changing the decision threshold. The bigger issue is that the confidence scores reported by MLLMs are not reliable. For all four MLLMs, higher reported confidence actually associated with lower accuracy (see Figure~\ref{fig:f13b}). Most MLLMs place their confidence near the top of the scale. Between 87\,\% of GPT-5.1 responses and 100\,\% of Gemini-3-Pro and Qwen3-VL-235B responses fall at confidence levels 6 or 7. The average reported confidence ranges from 6.4 to 6.8, compared to 5.18 for humans. For three models, this high confidence is clearly misleading. GPT-5.1 drops from 65\,\% accuracy at confidence level 5 ($n=156$) to 29\,\% at level 7 ($n=776$). Gemini-3-Pro drops from 91\,\% at level 6 to 51\,\% at level 7. Qwen3-VL falls from 66\,\% to 21\,\%. Gemini-3-Flash shows a milder version of the same issue, with accuracy decreasing from 72\,\% to 55\,\%, but without fully reversing the trend. The models also almost never abstain. Three of the four models never return \textit{Unsure}, and GPT-5.1 does so in only 9 out of 1,400 trials (0.6\,\%), all on synthetic images. This goes in the opposite direction from how humans hedge (see Section~\ref{sec:human-calibrated})

The calibration results tell the same story. Errors range from 0.32 for Gemini-3-Flash to 0.54 for GPT-5.1 and Qwen3-VL-235B, compared to 0.20 for humans. In other words, MLLM confidence is much less well aligned with actual accuracy. At the same time, the models seem to have more useful signal than their reported confidence suggests. If we ignore the final REAL/SYNTHETIC label and instead treat each (choice, confidence) pair as a single “looks synthetic” score, we can compare how often a synthetic image is ranked above a real one. By that measure, three of the four MLLMs outperform humans. Gemini-3-Flash gets the ordering right 81\,\% of the time, Gemini-3-Pro 78\,\%, and GPT-5.1 75\,\%, compared to 69\,\% for humans. Only Qwen3-VL falls below human performance, at 61\,\%. So the information is there inside the models—but the simple 7-point confidence scale that users see, clustered near the top and often pointing in the wrong direction, doesn’t make that information usable.

\paragraph{Key takeaways.} Frontier MLLMs possess better actual detection capabilities than they appear to. They tend to be overconfident and often point in the wrong direction. Furthermore, the models almost never select the \textit{unsure} option. In everyday consumer scenarios, these models are safe because they rarely confidently flag a real image as fake. However, in the field of evidence authentication, this decision-making pattern can lead to the mass failure to detect forgeries.

\subsection{How specialized detectors fail}
\label{sec:specialized-detectors}

The MLLMs we test in Section~\ref{sec:mllm-modes} are general-purpose vision-language systems, not detectors trained specifically on AI-generated images. We evaluated three specialized detectors on the same SLED-1400 stimuli, using the same balanced-accuracy framework.

Table~\ref{tab:specialized-detectors} shows the detection rates for each generator. Overall, these specialized detectors do not close the gap. They perform well on real images, with near-perfect specificity ($\geq 97.5\,\%$), but they struggle to detect synthetic images from the hard-regime generators. UnivFD detects none of the 1{,}200 synthetic images. CNNDetect detects only 1.2\,\%. ViT-Det performs best among the three, but still detects only 9.2\,\%, far below the four-MLLM average of 44.1\,\% and the human aggregate of 65.4\,\%. Given the weak performance of these specialized detectors, we do not include them in the later discussion of the hybrid framework (\ref{sec:disc-hybrid}).

\begin{table}[!htbp]
  \centering
  \small
  \setlength{\tabcolsep}{5pt}
  \renewcommand{\arraystretch}{1.15}
  \begin{tabular}{lccccccc}
    \toprule
    \textbf{Rater} & \textbf{Synth} & \textbf{Flux-2} & \textbf{Gem-2.5} & \textbf{Gem-3-Pro} & \textbf{GPT-1.5} & \textbf{Huny-3} & \textbf{Qwen} \\
    \midrule
    Human  & 0.654 & 0.510 & 0.624 & 0.485 & 0.525 & 0.831 & 0.930 \\
    MLLM     & 0.441 & 0.198 & 0.403 & 0.059 & 0.148 & 0.864 & 0.976 \\
    \midrule
    UnivFD           & 0.000 & 0.000 & 0.000 & 0.000 & 0.000 & 0.000 & 0.000 \\
    CNNDetect        & 0.012 & 0.000 & 0.005 & 0.000 & 0.020 & 0.045 & 0.000 \\
    ViT-Det          & 0.092 & 0.040 & 0.080 & 0.110 & 0.075 & 0.105 & 0.145 \\
    \bottomrule
  \end{tabular}
  \caption{Synthetic-image detection rate by rater and by generator, on SLED-1400.}
  \label{tab:specialized-detectors}
\end{table}

\subsection{Can Their Failures be Combined?}
\label{sec:combination}

Humans and MLLMs fail in very different ways, as the analysis above shows. We now ask whether we can combine the two to achieve higher accuracy.

\subsubsection{Different errors across rater types}
\label{sec:combination-orthogonal}

\begin{figure}[!htbp]
  \centering
  \includegraphics[width=0.65\linewidth]{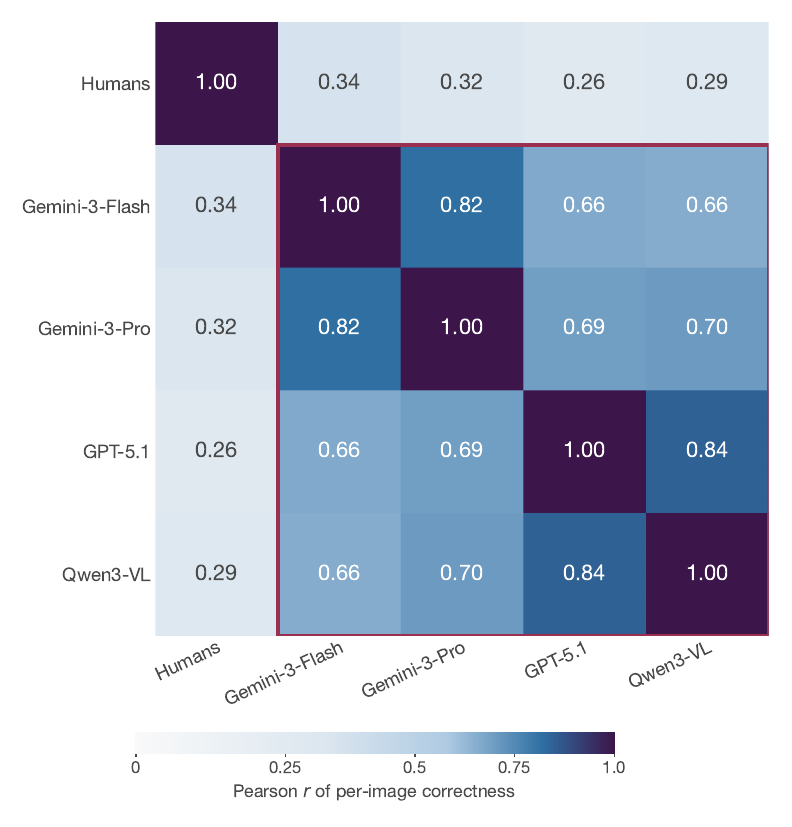}
  \caption{Pearson correlation between each pair of raters, computed across the 999 images that had at least one decided human judgement and a decided judgement from every MLLM (UNSURE responses excluded).}
  \label{fig:f10-corr}
\end{figure}

Figure~\ref{fig:f10-corr} shows how often each pair of raters was right on the same images. For each image we record whether each rater got it right, then take the Pearson correlation between every pair of raters acorss all images. Pearson $r$ runs from $0$ (no relationship between the two raters' correctness) to $+1$ (the two raters were right and wrong on the same images). The four MLLMs are strongly correlated with each other ($r = 0.66$--$0.84$, the red box in the figure). They tend to be right on the same images and wrong on the same images. Humans correlate only weakly with any single MLLM ($r = 0.26$--$0.34$).

\subsubsection{Combining MLLMs alone does not replace human input}
\label{sec:combination-no-substitute}

An ensemble of several MLLMs is a natural alternative to a human-MLLM hybrid. Our data show that it does not work. Of the 1{,}191 synthetic images on which all four MLLMs returned a decided answer, 38.8\,\% were missed by all four MLLMs and 27.4\,\% were caught by all four (Table~\ref{tab:ensemble}). If the four MLLMs were truly independent, those percentages would be very different. The average MLLM catches 44\,\% of synthetic images. If four such detectors were truly independent, the chance that all four miss the same image would be roughly $0.56^{4} \approx 9.9\,\%$, and the chance that all four catch it would be roughly $0.44^{4} \approx 3.7\,\%$. What we actually see is about four times the first number and seven times the second. The four MLLMs are highly correlated, so combining them does not buy four detectors' worth of coverage. Recovering the different error sets between humans and MLLMs (Section~\ref{sec:combination-orthogonal}) requires a human in the loop.

\begin{table}[!htbp]
  \centering
  \small
  \setlength{\tabcolsep}{8pt}
  \renewcommand{\arraystretch}{1.15}
  \begin{tabular}{lcc}
    \toprule
    \shortstack{\textbf{Number of MLLMs (of 4)}\\\textbf{that caught the image}} & \textbf{Empirical} & \shortstack{\textbf{Binomial baseline}\\\textbf{(if independent)}} \\
    \midrule
    0 (missed by all)  & \textbf{38.8\,\%} (462) & 9.9\,\% (118) \\
    1                  & 9.9\,\% (118)           & 31.0\,\% (369) \\
    2                  & 15.6\,\% (186)          & 36.4\,\% (433) \\
    3                  & 8.3\,\% (99)            & 19.0\,\% (226) \\
    4 (caught by all)  & \textbf{27.4\,\%} (326) & 3.7\,\% (44)  \\
    \bottomrule
  \end{tabular}
  \caption{Distribution of how many MLLMs (out of four) correctly flagged each synthetic image, on the 1{,}191 synthetic images on which all four MLLMs returned a decided verdict (UNSURE responses excluded).}
  \label{tab:ensemble}
\end{table}

\paragraph{Key takeaways.} Humans and MLLMs make largely different mistakes. The per-image correctness correlation and verdict agreement both place humans far outside the tight four-MLLM block, and the two rater types abstain on opposite kinds of image. Combining MLLMs alone does not recover that difference, because the four MLLMs already largely agree on their errors. This is the empirical basis for the tiered hybrid-verification proposal in Section~\ref{sec:disc-hybrid}.

\subsection{What raters say when they judge}
\label{sec:reason-patterns}

Sections~\ref{sec:human-modes}--\ref{sec:combination} describe what humans and MLLMs decide and how often those decisions are correct. They do not address why. The 1{,}574 human reasons and 5{,}599 MLLM reasons, coded under the procedure of Section~\ref{sec:reason-coding}, give a direct answer. \textit{All human reason quotes that follow are English translations, 98\,\% of human reasons were originally written in Chinese.}

\subsubsection{The ``no-anomaly'' default}
\label{sec:reason-template}

When a human or MLLM judged an image to be real, what did it say to justify the decision? In 96\,\% of GPT-5.1's \textit{Real} reasons the model used template phrasing such as ``no AI artifacts visible'', ``textures look natural'', or ``consistent with a real photograph'' rather than naming a specific feature of the image. Qwen3-VL-235B used the same template in 79\,\% of its \textit{Real} reasons; Gemini-3-Pro and Gemini-3-Flash in 45\,\% each; human participants in only 14\,\%. A typical GPT-5.1 reason on an image it wrongly called \textit{Real} reads:
\begin{quote}
``Natural lighting, depth of field, food textures and crumbs all resemble a typical real photograph without visible AI artifacts.''
\end{quote}
Every sentence names a general feature that could apply to almost any indoor photo, and none identifies anything specific to the image being judged.

The same template recurs almost word-for-word across Qwen3-VL's wrong-\textit{Real} reasons, with only the noun slots changing from image to image, for example:
\begin{quote}
``The image looks like a real photo with natural lighting and shadows, consistent textures on the shoes and box, and no obvious AI artifacts or distortions.''
\end{quote}
Gemini-3-Flash produces the same kind of phrasing on images it is confident about, with a slightly different surface vocabulary:
\begin{quote}
``The image shows natural fabric textures, realistic folding patterns, and a physically consistent metal hanger with appropriate shadows against the wall.''
\end{quote}
Gemini-3-Pro relies on templates less often than the other three. When it does, the template typically appears as a closing clause after one or more concrete observations:
\begin{quote}
``The image displays realistic physics in the car damage (torn plastic, crumpled metal, exposed radiator) and authentic environmental details like dust on the windshield and asphalt texture. The `Ford' logo is perfectly legible, and lighting/reflections are consistent with a real photograph.''
\end{quote}

The ranking across raters matches the accuracy ranking from Section~\ref{sec:mllm-overall} exactly. The two MLLMs with the lowest accuracy (GPT-5.1 41.9\,\%, Qwen3-VL 42.1\,\%) are the two that rely on templates most. The two with the highest accuracy (Gemini-3-Pro 60.4\,\%, Gemini-3-Flash 63.8\,\%) use templates less than half the time. When an MLLM defaults to listing what is absent rather than pointing to what is present, it tends to default to \textit{Real}, and on a corpus that is six-sevenths synthetic ($1{,}200/1{,}400$), defaulting to \textit{Real} is the main route to error.

\subsubsection{Different reasoning vocabularies}
\label{sec:reason-vocab}

Two coded categories appeared in roughly one in five human reasons and almost never appeared in any MLLM reason. The first is intuition. Human participants often admitted they could not give a specific reason:
\begin{quote}
``looks real, but feels off''\\
``nothing obviously wrong, but it doesn't quite feel real''
\end{quote}
Across all four MLLMs combined, less than one percent included similar free-text reasons. The second is everyday experience. Humans make judgments based on lived knowledge:
\begin{quote}
``I have seen this happen when doing laundry''\\
``this kind of plant decay is very common in daily life''\\
``what drink costs that much?''
\end{quote}
MLLMs almost never used reasoning of this kind, even though their pre-training corpus contains the same background knowledge. 

The cost of this difference appears on images where the visual surface is photorealistic but the situation it depicts is not. On a synthetic image in the Food Spoilage category, showing burnt toast plated next to freshly diced raw tomatoes, a human participant rejected it on pragmatic grounds:
\begin{quote}
``Burnt food can't be eaten, and nobody would plate it together with normal tomatoes.''
\end{quote}
All four MLLMs called the same image \textit{Real} with confidence 6 or 7. A representative GPT-5.1 reason reads:
\begin{quote}
``The food, plate, and background textures, lighting, and small imperfections all appear naturally photographic, with consistent detail and depth that resemble a real camera capture rather than AI artifacts.''
\end{quote}
The MLLMs analysed what they saw correctly. They did not ask whether what they saw was the kind of scene a person would actually photograph.

The Section~\ref{sec:combination-orthogonal} finding that human and MLLM errors are largely independent is grounded in this difference. Humans and MLLMs attend to different things and use different vocabulary when asked to explain themselves.

\subsubsection{Why humans fail on real receipts}
\label{sec:reason-receipts}

Here, we example the category where humans performed worst. For real receipts and lists, human wrongly called then synthetic in 51.3\,\% of the case (Section~\ref{sec:human-axes}). Given that this category is primarily text-related, the most intuitive guess is that humans overlook these textual clues. However, the data shows that humans do take anomalous text and symbols into account in their judgments (63\,\%).

When humans correctly accepted a real receipt, 52\,\% of their reasons used a brief positive template:
\begin{quote}
``looks like a real receipt''\\
``details are real''\\
``matches what real photographs look like''
\end{quote}
When humans wrongly rejected a real receipt, that template appeared in 0 of 46 reasons. The wrong-rejection reasons pointed instead at creases, structural details, and intuition:
\begin{quote}
``the creased and blurred parts do not join up naturally''\\
``the receipt is warped, not a clean rectangle''\\
``the text alignment and formatting are flawed''
\end{quote}

Creases, warping, and alignment irregularities are the legitimate imperfections of thermal-printed paper. The cause is seeing too much, not missing text cues. Telling raters to ``watch for text and symbol artifacts'' will not fix the failure, because they already do. The right intervention is calibration training that teaches them to distinguish legitimate photographic imperfections from genuine AI artifacts. The opposite failure appears on Electronics \& appliance damage, where humans miss 43.1\,\% of synthetic exhibits (Table~\ref{tab:f12-fp}).

\paragraph{Key takeaways.} The reason coding gives a concrete cause for each of the three quantitative findings of Sections~\ref{sec:human-modes}--\ref{sec:combination}. The MLLMs' templated phrasing on \textit{Real} verdicts is the linguistic side of their conservative decision rule, and the more a model relies on those templates the more often it is wrong. The gap between human and MLLM error sets is grounded in different reasoning vocabularies. Humans draw on intuition and everyday experience, while the MLLMs almost never do. Human's low performance on real receipts is a problem of seeing too much.

\section{Discussion}
\label{sec:discussion}

So, back to our title, can you trust what you see? Our results show a difficult detection landscape. Neither humans nor frontier MLLMs can serve as reliable standalone authenticators of visual evidence across settings. On images generated by Gemini-3-Pro-Image and Flux-2-Max, our 136 lay participants detect synthetic content at 48.5\,\% and 51.0\,\%, which is statistically indistinguishable from coin-flip. And three of the four frontier MLLMs we tested detect then at or below 5\,\%.

Earlier work in this space framed visual indistinguishability as a future scenario to prepare for \citep{delfino_deepfakes_2023_2}. Our data show it is now the current state at the frontier. We must respond by means of procedural and technical measures.

\subsection{Two failure mechanisms converge on the same conclusion}
\label{sec:disc-mechanisms}

Frontier image generators have been tuned to remove the features humans use to spot AI-generated images (unnatural lighting, illogical structures, and similar visual cues). The generators will keep improving. Human discrimination may improve too, slowly, with experience. But we are past the point where untrained viewers can reliably tell real images from AI-generated ones.

MLLMs fail differently. Their perception is in fact sharper than the average person's (Section~\ref{sec:mllm-overall}). They fail because they apply a very conservative decision rule. Before labelling an image SYNTHETIC, they require strong evidence. This rule keeps them safe in everyday use, but it is the wrong rule for evidence authentication.

This MLLM behavior likely comes from how the models are trained. Alignment training might penalizes a model that confidently calls a real image fake. The training data might also contains relatively few images from the newest generation of generators. Hedging is also limited. Three of the four models we tested never returned \textit{Unsure}, and the one that did used it just 9 times in 1{,}400 trials.

\subsection{No single technical fix closes the gap}
\label{sec:disc-no-fix}

Given that humans cannot reliably identify frontier-generated images, can we find a technical fix?

The first option is to combine several MLLMs into an ensemble. Our data show this helps very little. The four MLLMs we tested are highly correlated with each other (Section~\ref{sec:combination-orthogonal}), even though they come from different developers. They tend to be right on the same images and wrong on the same images. Nor do specialized detectors solve the problem. As shown in Section~\ref{sec:specialized-detectors}, the three specialized detectors we test are unreliable in this setting, and all three perform worse than the four general-purpose MLLMs.

The second fix is to require generators to mark every image with a watermark. The watermark can be a hidden signature in the pixel data or a cryptographic record bundled with the file, and any verifier can then check whether the image came from a generator. Google SynthID \citep{gowal2025synthid} and the C2PA Content Credentials standard\footnote{https://c2pa.org/} are two notable examples. The current coverage is partial. Not every commercial generator adopts the standard, and open-weight generators that anyone can run on a local machine skip the watermark entirely. Even the watermarks that do get added can be removed by a simple workflow. Generate the image with a strong generator, then pass it through a local image-to-image model that repaints it slightly. The result looks the same to the eye, but the watermark is gone.

Watermark removal is not the focus of this paper. We bring it up to show that no single technical fix currently closes the detection gap described in Section~\ref{sec:disc-mechanisms}.

\subsection{Toward a hybrid framework}
\label{sec:disc-hybrid}

Combining humans and MLLMs could improve accuracy because their errors are not the same (Section~\ref{sec:combination-orthogonal}). Humans and any single MLLM agree on which images are easy and which are hard only weakly ($r = 0.26$--$0.34$). The four MLLMs agree with each other much more strongly ($r = 0.66$--$0.84$). The two rater types rescue each other, but not equally. Humans recover a wrong MLLM majority on 18.8\,\% of images, and MLLMs recover a wrong human majority on 10.4\,\%. 

The combination still leaves some images unsolved. On 15.2\,\% of images both rater types get the wrong answer, and no method that relies on looking at the image can authenticate these. The hybrid is therefore better than any single rater type, but it does not solve the underlying problem of detecting frontier-generated images.

\section{Limitations}
\label{sec:limitations}

Several limitations are worth keeping in mind when interpreting these results.

\paragraph{Sample composition.} Our sample of 136 lay adults was recruited online without forensic, photographic, or legal training. The 64.8\,\% aggregate accuracy is a baseline for the lay population that evaluates visual evidence in high-volume, low-value civil disputes (consumers, merchants, platform moderators, claimants), not for judges, practicing attorneys, jurors, or trained forensic examiners.

\paragraph{Experimental ecology.} The web-based, self-paced setting removes the case files, testimony, corroborating exhibits, chain of custody, and party history on which courtroom authentication usually relies. They were also explicitly told that some images were AI-generated. In real-world settings, fabrication is usually just a possibility raised by one side.

\paragraph{Generator coverage and currency.} The six generators we tested span leading closed-source, open-weight, U.S., and Chinese systems as of late 2025 / early 2026, but not the full landscape. Newer models and domain-specific fine-tuning will likely change the exact performance numbers within few weeks. What is more likely to hold up over time is the broader pattern we observe, especially the detection turning point (see \ref{sec:human-axes}), rather than the specific results tied to individual models.

\paragraph{MLLM cross-family confound.} Three of the four MLLM evaluators we use come from the same model families as at least one of the generators. When we exclude same-family outputs, the main findings actually become stronger. Still, this does not rule out more subtle effects linked to shared architecture.

\paragraph{Scope of evidence categories.} Our ten categories focus on object-based photographic evidence commonly seen in consumer disputes, insurance claims, and product liability cases. They do not include other important types of evidence, such as criminal case materials (e.g., crime scene images, injury documentation, surveillance footage), technical expert evidence (e.g., forensic microscopy, satellite imagery), or biometric data (e.g., face recognition, gait analysis). While we expect the general patterns to extend to these areas, the specific numbers reported here should not be taken as directly transferable.

\paragraph{Reason coder reliability.} The qualitative analysis in Section~\ref{sec:reason-patterns} relies on a single LLM (Claude Sonnet 4.6) to code all 7,173 reasons. Because this coder is itself an MLLM, it may share some of the same blind spots as the models it is evaluating.

\section{Conclusion}
\label{sec:conclusion}

This study provides the first systematic empirical comparison of human and AI capabilities for detecting AI-generated legal visual evidence. More specifically, we investigated the extent to which humans and state-of-the-art MLLMs can reliably distinguish between real and fake visual evidence in legal contexts. Using a self-built dataset of 200 authentic evidence images and 1{,}200 synthetic counterparts generated by six leading generation models, we evaluated the detection performance of human participants and four MLLMs under controlled conditions.

The short answer is that neither group is reliable enough for the legal setting. Humans do slightly better than a coin flip. MLLMs are excellent at confirming that real images are real, but they keep missing sophisticated fakes. The generator used to create the image largely determines whether anyone can spot the fake at all. Humans and MLLMs often identify different issues, compensating to some extent for each other's blind spots — though they cannot fully cover all of them. Current image detection efforts remain fragmented, with no single method capable of handling the task on its own. This suggests that we need to adopt hybrid verification methods and seriously rethink how courts handle visual evidence.

These findings matter beyond the courtroom too. As synthetic images get better, visual evidence will likely shift from something courts presume to be trustworthy to something they treat as inherently open to challenge. That changes the game for procedural fairness, burdens of proof, and litigation tactics, since parties will increasingly be able to dispute or exploit doubts about whether an image is real. It also means technical infrastructure like provenance tracking, secure capture pipelines, and standardized verification protocols will need to become core parts of how evidence works.

At the same time, the ``arms race'' between generation and detection technologies shows no signs of slowing. What we measured is a snapshot of where things stand in 2026, and generative models will keep improving. This dynamic landscape calls for ongoing empirical evaluation and adaptive policy responses. In this context, the path forward is unlikely to lie in choosing between human and AI judgment. It's combining them in structured ways that play to each side's strengths and compensate for each side's weaknesses. Getting a clear picture of those strengths and weaknesses is the first step toward building safeguards that can protect the integrity of legal proceedings at a time when seeing something is no longer enough reason to believe it.

\section*{Statements and Declarations}

\subsection*{Funding}
Jinzhe Tan, Ali Ekber Cinar, and Karim Benyekhlef would like to thank the Cyberjustice Laboratory at
Université de Montréal, the LexUM Chair on Legal Information for their support of this research.

\subsection*{Data Availability}
The full SLED-1400 dataset (authentic images, AI-generated counterparts, prompt-generation outputs), the anonymized human response logs, and all MLLM response logs will be made publicly available in an open repository upon acceptance of this article.

\subsection*{Code Availability}
The evaluation pipeline, web-based experimental platform source code, and analysis scripts used to generate the results reported in this article will be released under an open-source licence upon acceptance, alongside the dataset.

\bibliographystyle{sn-basic}
\bibliography{bibfile}

\end{document}